\documentclass[11pt,a4paper]{article}
\usepackage[T1]{fontenc}
\usepackage[utf8]{inputenc}
\usepackage{lmodern}
\usepackage[margin=0.85in]{geometry}
\usepackage{amsmath,amssymb,booktabs,graphicx,array}
\usepackage[numbers,sort&compress]{natbib}
\usepackage{xcolor}
\usepackage[colorlinks=true,linkcolor=blue!45!black,citecolor=blue!45!black,urlcolor=blue!45!black]{hyperref}
\usepackage{url}
\graphicspath{{figures/}}
\hypersetup{pdftitle={A First Glance at Jev for Network Traffic Classification},pdfauthor={Shenghe Xu and Lifan Mei}}
\title{A First Glance at Jev for Network Traffic Classification:\\Accuracy, Processing Time, and Cost}
\author{Shenghe Xu\thanks{Work performed outside the author's role at Amazon. Email: \href{mailto:shenghexu@gmail.com}{\texttt{shenghexu@gmail.com}}.}\\Amazon.com, Inc.
\and Lifan Mei\thanks{Corresponding author: \href{mailto:lifan.mei@xjtlu.edu.cn}{\texttt{lifan.mei@xjtlu.edu.cn}}.}\\Xi'an Jiaotong-Liverpool University}
\date{September 30, 2026}
\begin{document}
\maketitle
\begin{abstract}
% GENERATED from chapters/; edit Markdown and run build_manuscript.py.

We evaluate Jev on ten dataset-defined application labels in CESNET-QUICEXT-25 using only the first ten packets' sizes, directions, and inter-packet times. To the best of our knowledge, this is the first empirical study of general-purpose decision models, represented here by Jev, for application classification of network flows. Across 52,000 records from 26 collection weeks following the training period, 40 fixed labeled examples raise Jev's accuracy from 9.80\% to 28.42\%. Random Forest and Extra Trees trained on 8,000 records achieve 69.95\% and 66.80\% and outperform Jev in every week. Increasing Jev's context to 150 examples yields 34.50\% on the first test week. On a paired 100-record subset, Jev with 40 examples achieves 29\% accuracy at a median request time of 0.750 s, versus 37\% and 6.036 s for the generative language model OpenAI GPT-5.6 Sol with high reasoning effort through Azure; Jev also incurs lower API charges. The paired subset does not establish an accuracy advantage for either service, and the timing reflects different service configurations. Thus, labeled examples substantially improve Jev, but the tested Jev configurations remain less accurate than trained tree ensembles; unequal supervision budgets and fixed configurations prevent attributing the gap to a single cause.

\end{abstract}

% GENERATED from chapters/; edit Markdown and run build_manuscript.py.
\section{Introduction}

The appeal of general-purpose language models, made familiar through ChatGPT, extends beyond producing text. Instructions and a few examples can describe a task to an existing model without updating its weights \cite{brown2020fewshot}. For networking, this suggests a practical possibility: using a shared model service to interpret measurements and supply decisions, reducing the need to build a separate predictor for each task. Research on adapting language models to networking has begun to explore this possibility \cite{wu2024netllm}. Whether that flexibility translates into useful decisions depends on the observations available, the supervision supplied, and the operational cost of each prediction.

Early traffic classification brings these requirements together. Network monitoring and service management need to associate flows with applications, while encryption limits directly readable content and waiting for a complete flow delays identification. Packet sizes, directions, and timings provide a compact alternative to application content \cite{cesnetdataset,datazoofeatures}. A classifier trained on these measurements must also confront traffic collected after its training period. This creates a reason to investigate adaptation through a small labeled context: examples can be supplied to an existing service without fitting new weights. Such flexibility would be useful if it preserved sufficient accuracy at an acceptable processing time and cost.

This problem is particularly informative for evaluating that proposition. The observed state is short, the requested answer is one of a finite set of application names, and labeled examples can demonstrate the intended mapping. These properties make the task straightforward to express through an instruction-based interface. Its statistical difficulty remains substantial: anonymous byte counts and time intervals do not explain their relationship to an application name. Removing domains and other identifiers therefore exposes the question of whether a pretrained service can use numerical traffic evidence, rather than infer a label from an explicit semantic cue. Suitability for expressing a task and competence at solving it must be evaluated separately.

Jev offers a direct interface for such decisions: a caller supplies a state, a typed question, and candidate answers \cite{jevdocs}. Its provider emphasizes fast decisions that software can consume \cite{jevlaunch}. General-purpose GPT APIs also support structured outputs \cite{openaistructured}, so output format alone does not distinguish useful classifiers. The empirical question is whether Jev offers a favorable combination of accuracy, processing time, and cost relative to both a generative service and conventional supervised models. We study Jev as an observable service, without inferring its internal architecture from its interface. Our generative comparison uses the GPT-5.6 Sol API through Azure.

We compare all main methods on the same numerical prefixes from ten fixed applications in CESNET-QUICEXT-25. Random Forest (RF) and Extra Trees (ET) are fitted on 8,000 records from collection weeks 1-4 (June 1-28, 2024), then held fixed alongside Jev's context for 52,000 records from weeks 5-30 (June 29-December 27, 2024). Extensions on week 5 (June 29-July 5, 2024) examine a larger context and a paired generative-service comparison. The application is unknown for each query, but the candidate classes are fixed: the study concerns later-period classification within a closed set, rather than recognition of unseen classes or a presumed out-of-distribution shift.

To the best of our knowledge, this is the first empirical study of general-purpose decision models, represented here by Jev, for application classification of network flows. Our longitudinal evaluation uses only the first ten QUIC packets of each flow. A related Jev IDS project studies intrusion and attack-category detection from network-flow features \cite{jevids2026}; our question is which application generated an early QUIC packet prefix. The study connects three measurements: longitudinal classification quality, improvement from labeled context, and the processing time and API cost of the specified deployments. Context improves Jev substantially, yet the trained ensembles retain an accuracy advantage in every test week. On the paired subset, Jev has shorter observed requests and lower API charges than the tested high-reasoning Sol setting. Together, these results clarify what adapting a service through examples achieves in this task, and where the remaining accuracy gap limits its practical appeal.

% GENERATED from chapters/; edit Markdown and run build_manuscript.py.
\section{Background and Related Work}

\textbf{Generative and typed-decision interfaces.} A GPT-based classifier can receive a task description, measurements, candidate labels, and demonstrations, then return a label in a structured response. OpenAI's Structured Outputs supports schema-constrained responses, including enumerated choices; JSON mode ensures valid JSON but does not itself enforce a particular schema \cite{openaistructured}. Our Sol configuration uses JSON mode with local label validation. Jev instead exposes typed questions directly, including a choice over supplied candidates \cite{jevdocs}. The relevant distinction is how each service exposes and processes a decision, not whether software can consume its output. We compare the recorded configurations; neither their interfaces nor their names establish architectural speed, correctness, or calibration.

\textbf{Why evaluate language models on traffic measurements?} In-context learning provides a way to communicate a task through labeled examples without parameter updates \cite{brown2020fewshot}. Early traffic classification supplies a compact input and a well-defined output, making it a tractable setting in which to test that mode of adaptation. However, successful language understanding does not imply a learned correspondence between packet measurements and application identities. NetLLM addresses the gap between networking inputs and language models through modality encoding, task-specific output heads, and parameter-efficient adaptation \cite{wu2024netllm}. Its results motivate studying adaptation; they do not establish that serializing numerical prefixes for an unadapted service is sufficient. TabPFN, a foundation model designed for tabular prediction, further illustrates the importance of the model's training and representation \cite{hollmann2025tabpfn}. We evaluate the simpler service-based route, without extending our findings to these specialized approaches.

\textbf{Traffic representations and evaluation scope.} CESNET-QUICEXT-25 provides longitudinal QUIC measurements \cite{cesnetdataset,muecke2025quic}, including per-packet timing, direction, and size \cite{datazoofeatures}. Our short numerical prefix makes the observed information comparable across tree ensembles and hosted services. ET-BERT instead learns contextual datagram representations through pretraining and task adaptation \cite{lin2022etbert}. Its inputs and supervision differ, so its published scores are not numerical baselines for this experiment. Here ET means Extra Trees. This study compares methods using the same short numerical prefixes across later collection weeks under the recorded training and service configurations, rather than surveying traffic-classification methods exhaustively.

\textbf{Evidence about Jev.} Public documentation describes Jev's interface and acknowledges uneven capabilities, including numerical tasks \cite{jevdocs,jevlimits}. A public Jev IDS project uses NSL-KDD flow features to classify traffic as normal or malicious and to identify attack categories \cite{jevids2026}. Our task instead identifies one of ten application/service labels from only the first ten QUIC packets, with fixed inputs evaluated across 26 later collection weeks. A recent survey maps Jev projects across domains rather than benchmarking network application classification \cite{ling2026jevwild}. Our 52,000-record comparison measures accuracy, request time, and API charges for the specified configurations; a separate paired subset compares Jev with a generative service. We keep three properties separate: returning an admissible value, selecting the correct application, and delivering that answer within a useful time and cost. Only the observed classification and service behavior are tested here; weights, internal computation, and confidence calibration are outside the study.

% GENERATED from chapters/; edit Markdown and run build_manuscript.py.
\section{Experimental Design}

\subsection{Data and isolation}

We use a fixed archived revision of a public Parquet conversion of the CESNET DataZoo ORIG distribution \cite{mirror}. The 205 daily files cover nominal June 1-December 27, 2024 collection partitions and contain 110,511,231 records before filtering. Sizes and SHA-256 hashes match the archived mirror. Additional checks found no differences in packet sequences and timing for 14,750 first-day records against the official raw source, or in checked IDs, labels, and sequences of 8,000 previously obtained official HDF5 records. This verifies provenance, not every label's correctness.

Weeks are seven-day \textbf{daily-file collection partitions}, not asserted UTC flow-start intervals: a record's first timestamp can precede its filename date. Weeks 1-4 correspond to June 1-28; week 5 to June 29-July 5; week 30 to December 21-27. Missing dates, October 31 and December 11-14, are not replaced.

The fixed classes are \texttt{apple-privaterelay}, \texttt{facebook-graph}, \texttt{google-ads}, \texttt{google-gstatic}, \texttt{google-play}, \texttt{google-services}, \texttt{google-www}, \texttt{instagram}, \texttt{snapchat}, and \texttt{youtube}. Labels are inherited from the source; the conversion documents SNI-derived label provenance \cite{mirror}. Domain strings are not model inputs, although candidate class names are visible as the output vocabulary.

Records must have at least ten packets. Each input comprises ten triples: inter-packet time in milliseconds, direction encoded as -1/+1, and transport-payload size in bytes \cite{datazoofeatures}. Values are preserved without rescaling. RF/ET receive a fixed channel-major vector of 30 float32 values; services receive the same measurements as packet-major JSON triples with field descriptions. No test label, domain, address, port, or absolute time is supplied.

A fixed seeded hash of source ID (seed 2026092305) prioritizes eligible records, with at most 8,192 candidates per week. We retain the first 2,000 after excluding previously selected IDs and exact prefix fingerprints. This operates on raw-record priorities before deduplication, not uniform sampling over distinct prefix groups, and stops if the pool is insufficient. Selection is not class-balanced or prediction-dependent. All 60,000 selected training/test IDs and exact float32-prefix fingerprints are unique. This does not establish disjoint hosts, users, or underlying connections.

\subsection{Models and labeled context}

\textbf{RF and ET.} Random Forest \cite{breiman2001rf} and Extra Trees \cite{geurts2006et} each use 500 trees, square-root feature subsampling, Gini splitting, minimum leaf size one, unlimited depth, no class weighting, and seed 2026092305 in scikit-learn 1.7.2. They are fitted once on the same 8,000 records and reused without test-based tuning. Fitting uses four jobs; timed single-record prediction uses one job after warm-up. These are reference ensembles, not an exhaustive state-of-the-art suite. A constant reference chooses only the training-majority class, \texttt{google-ads} (1,402/8,000 training records); its full-set score is consolidated during manuscript preparation.

\textbf{Jev.} The requested OpenRouter model is \texttt{typesafe/jev-1.13}, checked against returned identifier \texttt{typesafe/jev-1.13-20260917}. One choice question requests exactly one application. Jev-0 and Jev-40 denote zero and 40 labeled examples, respectively. The 40-example state uses the lowest-ranked training record in each week/class cell: four examples per class, fixed throughout evaluation, with no weight updates or test feedback.

The 150-example extension retains the original 40 as its prefix and adds fixed-ranked training records to reach 15 per class and 37/37/38/38 per training week. It evaluates the original 2,000 week-5 records. One record checks context capacity and remains in the accuracy denominator; the remaining 1,999 use eight workers. Input counts are 23,614-23,635 tokens. These contexts are class-balanced, unlike the full training set.

\textbf{GPT-5.6 Sol.} The generative baseline is OpenAI GPT-5.6 Sol, accessed through Azure. A fixed salted SHA-256 ranking of week-5 sample IDs selects 100 records without consulting labels or predictions. Deployment \texttt{gpt-5.6-sol}, with returned identifier \texttt{gpt-5.6-sol-2026-07-09}, uses \texttt{high} reasoning effort, a 4,096-completion-token cap, non-streaming JSON output, and a 60 s request deadline. Sol-0 and Sol-40 denote this same configuration with zero or the original 40 examples. Inputs, candidates, and examples match Jev; the API/output wrapper differs. Serial requests alternate arm order by sample. Both services and RF/ET are compared on this identical subset. It is exploratory: week 5 had already been examined, and this is neither a blind new holdout nor an optimized LLM configuration search.

\subsection{Metrics and operational accounting}

Accuracy uses the fixed planned denominator. Macro-F1 averages the ten per-class F1 scores, assigning zero to undefined scores. Final coverage is complete: 52,000 per main arm, 2,000 for Jev-150, and 100 per Sol arm. The smaller studies overlap the main test set.

Local time encloses one warm-started \texttt{predict} call. Remote time includes client/worker communication, network, gateway, and service processing for a successful request. Neither includes observing the ten packets. Later-week ML batch times are not presented as single-flow latency; paired timing uses the original week-5 single-record measurements. Jev uses eight workers per main arm with partially overlapping execution; Sol is serial at different times over a different service path. Medians and 95th percentiles therefore compare deployments, not isolated architectures under matched load.

Interrupted runs preserve successful predictions and resume only missing queries with unchanged settings. Each query stops at its first valid answer, irrespective of correctness. Failures, retries, and unknown-charge attempts remain in accounting. Successful-attempt time excludes retry waits and does not measure first-attempt availability.

Jev cost comes from returned usage. Sol uses conservative bookkeeping rates of USD 6.25/M input and USD 30/M output tokens without cache discounts, not a verified invoice or current-tariff claim. Conservative totals additionally retain a 25\% allowance and failed/unknown-request reservations. Local RF/ET have no API fee; hardware, energy, and labor are unpriced. The September 23, 2026 runs used Windows 11, Python 3.12.14, NumPy 2.3.5, and an Intel family-6/model-158 CPU with 12 logical processors, without a local GPU. Frozen inputs, model identifiers, response hashes, and labels were locally cross-checked; aggregate data accompany this paper.

% GENERATED from chapters/; edit Markdown and run build_manuscript.py.
\section{Results}

\subsection{Full-set accuracy and prediction behavior}

\autoref{tab-full} reports the same 52,000 records. Forty examples improve Jev by 18.62 percentage points, but leave a 41.53-point gap to RF. Macro-F1 preserves the ordering among RF, ET, and the two Jev configurations. The training-majority constant reaches 22.79\% accuracy: Jev-40 exceeds it by 5.62 points and has substantially higher macro-F1, whereas zero-shot Jev is below it on both metrics.

\begin{table}[tbp]
\centering\small
\caption{Full test set, with 100\% final valid coverage for every method. Accuracy is percent; supervision budgets differ.}
\label{tab-full}
\begin{tabular}{@{}lrrr@{}}
\toprule
Method & Correct & Accuracy & Macro-F1 \\
\midrule
Training-majority constant & 11,852 & 22.79 & 0.0371 \\
RF (8,000 training) & 36,372 & 69.95 & 0.7155 \\
ET (8,000 training) & 34,736 & 66.80 & 0.6898 \\
Jev-0 & 5,095 & 9.80 & 0.0338 \\
Jev-40 & 14,776 & 28.42 & 0.3799 \\
\bottomrule
\end{tabular}
\end{table}

Zero-shot predictions concentrate on \texttt{google-www} (45,010), \texttt{youtube} (6,958), and \texttt{google-gstatic} (32), with no other predicted class. This is not uniform random guessing across balanced classes. Context helps unevenly: Jev-40 recalls 98.3\% of 815 \texttt{apple-privaterelay} records, close to RF's 99.4\%, but only 15.8\% of \texttt{google-ads} and 16.9\% of \texttt{google-play}, versus RF's 78.3\% and 74.2\%. These differences diagnose the aggregate result; favorable classes are not selected to replace it.

\subsection{Across collection weeks}

\begin{figure}[tbp]
\centering
\includegraphics[width=0.90\linewidth]{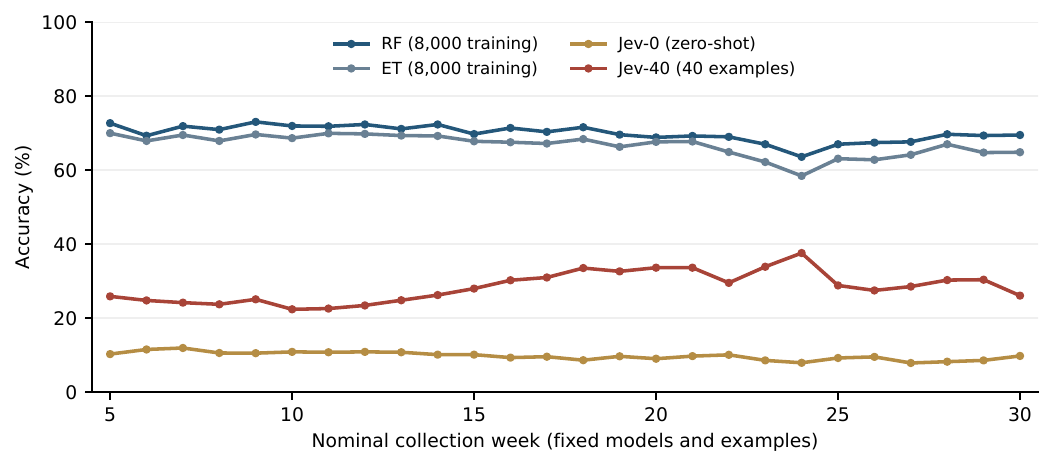}
\caption{Weekly accuracy on 2,000 identical test records per week, with fixed models and context. These are descriptive collection-week trajectories.}
\label{fig-weekly}
\end{figure}

RF and ET remain above Jev-40 in all 26 test weeks (\autoref{fig-weekly}). RF changes from 72.70\% in week 5 to 69.50\% in week 30, and ET from 70.00\% to 64.85\%. Both reach their minima in week 24 (63.60\% and 58.45\%) and then recover. Jev-40 peaks at 37.60\% in that week without overtaking either ensemble. There is no observed monotonic collapse or accuracy crossover. Changing class proportions, traffic behavior, and missing dates may contribute; chronological separation alone does not establish an out-of-distribution advantage.

\subsection{More labeled examples}

\begin{figure}[tbp]
\centering
\includegraphics[width=0.84\linewidth]{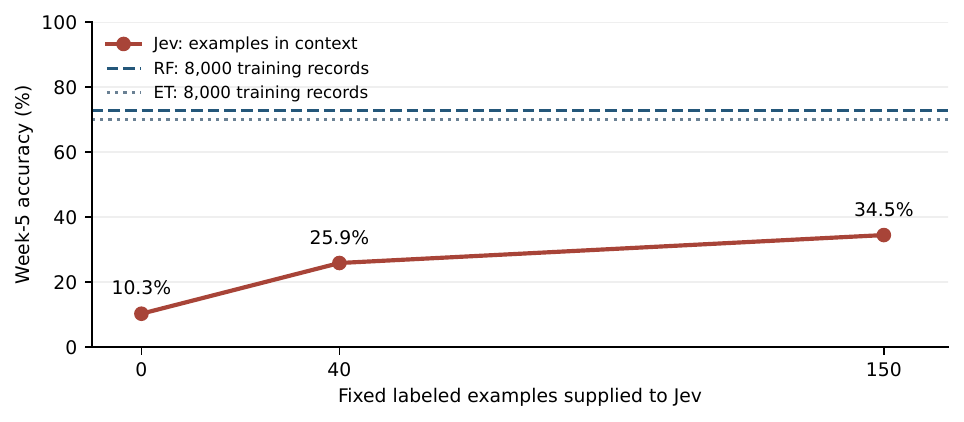}
\caption{Week-5 accuracy versus Jev context size on the same 2,000 records. The RF/ET reference lines use 8,000 training records, not equal label budgets.}
\label{fig-context}
\end{figure}

On the same 2,000 week-5 records, Jev with zero, 40, and 150 examples correctly classifies 206, 518, and 690 records, respectively: 10.30\%, 25.90\%, and 34.50\% (\autoref{fig-context}). The 150-example arm has macro-F1 0.4560 and remains 38.20 accuracy points below RF. Gains of 15.60 and 8.60 points show that this fixed context family helps, but three points from one week do not establish a scaling law. All 2,000 requests succeed; the 1,999-record eight-worker stage has a 0.910 s median. Its single-worker capacity probe is excluded only from that timing summary. No matched-40/150-example tree experiment has yet been run.

\subsection{Paired generative-service comparison}

\autoref{tab-paired} uses the same 100 week-5 records throughout. Sol-40 correctly classifies 37 versus Jev-40's 29. Sol alone is correct on 19 records, Jev alone on 11, and both on 18. An exploratory two-sided exact McNemar test gives p = 0.2005: this small sample does not establish stable superiority or equivalence, and shared collection conditions limit population-level interpretation. RF/ET achieve 65/70; their reversed ordering relative to the full set also cautions against replacing the main experiment with this subset.

\begin{table}[tbp]
\centering\small
\caption{Paired week-5 subset (100 records). Suffixes denote context examples. Sol is OpenAI GPT-5.6 Sol through Azure, with high reasoning effort. P50/P95 are successful prediction/request times in seconds. RF/ET are local; Jev/Sol have different concurrency and service paths.}
\label{tab-paired}
\begin{tabular}{@{}lrrr@{}}
\toprule
Method & Accuracy (\%) & P50 (s) & P95 (s) \\
\midrule
RF & 65 & 0.028 & 0.033 \\
ET & 70 & 0.027 & 0.034 \\
Jev-0 & 12 & 0.707 & 0.832 \\
Jev-40 & 29 & 0.750 & 0.877 \\
Sol-0 & 3 & 18.666 & 45.742 \\
Sol-40 & 37 & 6.036 & 37.791 \\
\bottomrule
\end{tabular}
\end{table}

\begin{figure}[tbp]
\centering
\includegraphics[width=0.90\linewidth]{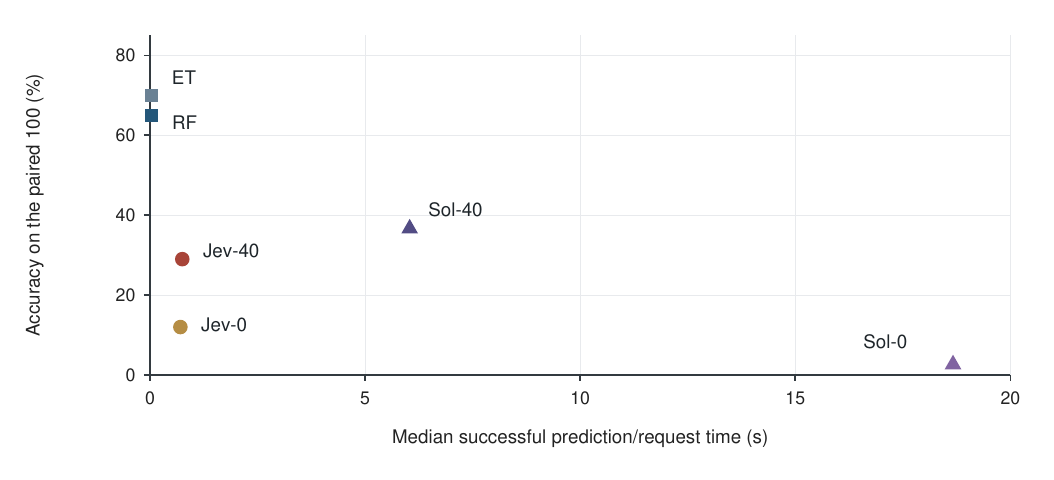}
\caption{Observed accuracy and median successful processing time on the paired 100 records. Squares are local ML, circles Jev, and triangles OpenAI GPT-5.6 Sol with high reasoning effort through Azure. Suffixes -0/-40 denote context examples. Timing boundaries differ.}
\label{fig-tradeoff}
\end{figure}

The 40-example request medians differ by 5.29 s, a ratio of about 8.0 (\autoref{fig-tradeoff}). This describes service configurations, not architectural speedup. Full-set Jev-0/40 medians are 0.716/0.773 s, with P95 of 0.832/0.911 s. Sol's usage records help interpret its longer response time: zero-shot median completion/reasoning counts are 1,049/1,024 tokens, versus 368.5/350.5 with context, despite prompt medians increasing from 260 to 5,676 tokens. Longer context coincides with less reasoning and shorter time, but does not causally isolate this mechanism. Lower reasoning effort was not tested.

\subsection{Failures and cost}

Sol obtains 200 valid labels over 211 attempts. Eleven failures comprise two client permission errors, seven deadlines, one invalid length-limited response, and one HTTP error. One query needs four attempts; its first valid response is incorrect and retained. Successful-request durations sum to 54.6 minutes, or 62.9 including failures, excluding inter-run gaps. These sums are not total experiment wall-clock time. Jev-0/40 record 52,008/52,020 started attempts, retaining terminal failures and 7/15 starts without terminal records after interruption. Unknown charges are not treated as zero.

\begin{table}[tbp]
\centering\small
\caption{Batch cost in USD. Jev known cost is API-reported; conservative totals include margins and failed/unknown reservations. Sol uses budget-rate estimates, not an invoice.}
\label{tab-cost}
\begin{tabular}{@{}lrrr@{}}
\toprule
Arm & Valid predictions & Jev known cost & Conservative cost \\
\midrule
Jev-0 & 52,000 & 1.217644 & 1.535495 \\
Jev-40 & 52,000 & 14.682004 & 18.386105 \\
Jev-150 & 2,000 & 1.984208 & 2.480260 \\
Sol two arms, incl. retries & 200 & -- & 14.013184 \\
Total & 106,200 & -- & 36.415045 \\
\bottomrule
\end{tabular}
\end{table}

The conservative batch total is USD 36.4150 (\autoref{tab-cost}), excluding earlier pilots and research. For 100 paired successful responses, Jev-0/40 report USD 0.002341/0.028234; Sol estimates are USD 3.927254/5.613564 before failures and allowances. These are service-price observations, not invoice-verified or energy-efficiency comparisons.

% GENERATED from chapters/; edit Markdown and run build_manuscript.py.
\section{Discussion and Limitations}

\textbf{What the accuracy gap means.} The task is easy to state but requires a mapping that its numerical inputs do not make explicit. Concentrated zero-shot predictions and gains from labeled context suggest that examples help supply this missing task information, although they do not reveal how either service reasons internally. Increasing Jev's context improves accuracy without closing the gap to the trained ensembles. The remaining gap may reflect the representation, example count and selection, service capability, or several of these together; this design cannot separate them. It does not imply that language models are unsuitable for language-rich logs, network intents, or other representations. RF/ET use 8,000 training records, whereas Jev receives 40/150 class-balanced examples and Sol at most 40; these results compare practical configurations, not intrinsic label efficiency. No matched-example tree control was run, so this study cannot estimate intrinsic label efficiency.

\textbf{What the service comparison means.} Jev's shorter requests and lower charges make its decision interface worth examining, but operational efficiency is useful only at an acceptable error rate. The paired subset does not resolve an accuracy advantage for either hosted service. Moreover, high-reasoning Sol is one configuration rather than an optimized speed baseline. API wrappers, concurrency, execution times, network paths, caching, and recovery differ, while local ML includes Python prediction overhead. We do not measure packet-observation time, feature extraction, live deadline satisfaction, or local Jev inference. The measured trade-off therefore concerns these deployments; it establishes neither real-time suitability nor an architectural speed advantage.

\textbf{Limits of the evidence.} One network, ten classes, one weekly selection, and one fixed context family limit generalization. Earlier exploratory work influenced scope, and week-5 extensions reuse examined data. Exact ID/prefix isolation does not exclude near duplicates or shared host/session conditions; hosted-model pretraining overlap is unknown. Labels are inherited from the dataset. Weekly trajectories describe later-period performance, but correlated flows and chronological separation alone cannot identify a causal drift mechanism or establish behavior in independent environments.

\textbf{Reproducibility and next steps.} The results apply to the recorded model identifiers and execution date; services and prices may change. Aggregate data and frozen-source hashes support the tables and figures, without constituting a public release of every raw response. Multiple example selections, independent networks, and lower-reasoning Sol settings could test how widely these observations hold. These extensions have not been run. Until then, the evidence supports a bounded conclusion: examples help adapt this decision service, while the accuracy needed to replace the tested supervised classifiers remains unachieved.

% GENERATED from chapters/; edit Markdown and run build_manuscript.py.
\section{Conclusion and Future Work}

Early traffic classification provides a concrete test of turning a pretrained service into a classifier through instructions and examples. On the ten-application numerical-prefix task studied here, labeled context improves Jev, including a further gain from 40 to 150 examples on week 5, but RF/ET trained on 8,000 records remain more accurate in every later collection week. Jev's shorter requests and lower charges relative to one high-reasoning Sol configuration reveal an operational trade-off, with the paired accuracy difference unresolved. In this setting, a convenient decision interface did not yield accuracy comparable to the trained ensembles.

Future work should test whether Jev can make better use of early-packet evidence through alternative representations and systematically selected examples. Matched-label-budget tree controls would clarify how much of the observed gap reflects supervision, while evaluation on independent networks and collection periods would test generality. Comparing service configurations under aligned timing and cost conditions would further establish where decision services may be useful for traffic classification. These directions explore Jev's potential without assuming an advantage that the present results do not show.

\bibliographystyle{unsrtnat}
\bibliography{references}
\end{document}